# Rule-based Knowledge Representation for Service Level Agreements

Doctoral Symposium of MATES'06

| Author | Supervisor |
| --- | --- |
| Dipl. Wirtsch.-Inf. Adrian Paschke | Prof. Dr. Martin Bichler |
| Internet Based Information System (IBIS) | Internet Based Information System (IBIS) |
| Technische Universität München | Technische Universität München |
| Adrian.Paschke@in.tum.de | bichler@in.tum.de |

**Abstract:** Automated management and monitoring of service contracts like Service Level Agreements (SLAs) or higher-level policies is vital for efficient and reliable distributed service-oriented architectures (SOA) with high quality of service (QoS) levels. IT service provider need to manage, execute and maintain thousands of SLAs for different customers and different types of services, which needs new levels of flexibility and automation not available with the current technology. I propose a novel rule-based knowledge representation (KR) for SLA rules and a respective rule-based service level management (RBSLM) framework. My rule-based approach based on logic programming provides several advantages including automated rule chaining allowing for compact knowledge representation and high levels of automation as well as flexibility to adapt to rapidly changing business requirements. Therewith, I address an urgent need service-oriented businesses do have nowadays which is to dynamically change their business and contractual logic in order to adapt to rapidly changing business environments and to overcome the restricting nature of slow change cycles.

## 1. State of Art, Problem Definition and Research Question

Service Level Agreements (SLAs) defining the performance criteria a provider promises to meet while delivering a service, are of growing commercial interest with a deep impact on the strategic and organisational processes. A well-defined and effective SLA correctly fulfils the expectations of all participants and defines the quality attributes and guarantees a service is required to process. It typically also sets out the remedial actions and any penalties that will take effect if performance falls below the promised service levels. During the monitoring and enforcement phase the agreed SLA rules will be used to detect violations to the promised service levels and to derive consequential activities in terms of actions, rights and obligations. They play a key role in metering, accounting and reporting and provide data for further refinement of SLAs and for optimizing service management on an operational, tactical and strategically level.

In practice, IT service providers need to manage, execute and maintain thousands of SLAs for different customers and different types of services in the upcoming service-oriented computing landscape. Commercial service level management tools such as IBM Tivoli, HP OpenView, CA Unicenter, BMC Patrol, and Microsoft Application Center store selected QoS parameters such as availability or response time as parameters in the application code or database tiers. This approach is restricted to simple, static rules with only a limited set of parameters. Existing SLA specification languages such as WSLA, WSML or WSOL are pure mark-up serialization languages which need specialized procedural interpreters, have a restricted expressiveness to describe complex decision and contractual logic in terms of rules and provide no capabilities to declaratively implement new functionalities. **As a result the upcoming service orientation based on services that are loosely coupled across heterogeneous, dynamic environments needs new ways of SLA representation with a high degree of agility and flexibility in order to efficiently manage, continuously monitor and enforce large amounts of complex and possibly distributed SLAs.**

I propose a novel declarative rule-based approach to SLA representation and management. Whereas, existing approaches to standardize SLAs are based on imperative procedural or simple propositional logic, I draw on logic programming (LP) and related knowledge representation (KR) concepts. LP allows for a compact representation of SLA rules and for automated rule chaining by resolution and variable unification, which alleviates the burden of having to implement extensive control flows as in imperative programming languages and allows for easy extensibility. However, further logical formalisms are needed for automated SLA management. I propose a novel, unique, highly expressive and efficient framework of adequate knowledge representation (KR) concepts. The underlying research question of this thesis is:

***Can advanced knowledge representation (KR) concepts using logic programming (LP) techniques and generic inference engines (rule engines) adapted to the SLA domain and will they enable an efficient, flexible and distributed management and enforcement of SLAs?***

## 2. Objectives and Contribution of the Work

The proposed research is on the interplay of common software engineering providing well-established development methodologies and imperative program and database solutions, on the one hand and artificial intelligence and knowledge engineering following the declarative rule based programming paradigm, on the other. A third important research dimension is added due to the inherent business-driven character of the application domain, the indispensable compliance with legal regulations and the deep impact on all operational, tactical and strategic IT service management (ITSM) processes. The first field refers to the adoption of successful SE methodologies, principles and techniques such as agile test-driven development, object oriented imperative programming and SE principles such as data abstraction or modularization. For the reason of acceptance, (re)usability with respect to existing expertise and tools, and interoperability in a distributed environment it is vital for a practical SLA management tool to be based on established SE methodologies and reuse existing development tools and existing operational systems such as network and system management tools, relational databases / data ware houses or middle ware applications. The legal and business driven aspects of SLAs induce high requirements regarding reliability, verifiability and traceability of derived conclusions (e.g. derived penalties) and triggered reactions which must also count in the legal sense and fulfil legal regulations and compliance rules. Moreover, service-oriented businesses face rapidly changing environments which amount for frequently adapting the business and contract logic. Obviously, this amounts for an interaction between the technical layer and the business layer as well as between various roles reaching from technicians, business practitioners to domain experts and executives. The objective for the proposed research is to adapt the values, principles and practices in the addressed fields and bring together the different requirements within one common framework. I argue that my rule-based SLA approach using logic programming and inference engines as the "common denominator" is well suited to be the methodical and technical core for this goal. Clearly, traditional logic programming and standard inference engines need to be adequately extended with further expressive logical formalisms and KR techniques to represent sophisticated contractual logic and incorporate external data and systems into declarative rule descriptions. This thesis contributes with a declarative, compact and highly expressive KR proposal for formalizing SLA specifications, which has the following properties:

- has a clear formal semantics,
- is computationally feasible even for larger SLAs and data sets
- reliable and traceable even in case of incomplete or contradicting knowledge
- provides support for validation and verification of contractual specifications
- is flexible in a way that allows to quickly alter the behaviour of the SLA system
- supports *programming* of arbitrary functionalities and decision procedures
- provides means for serialization and interchange based on a superimposed SLA mark-up language
- enables reuse and integration of external data, systems and object-oriented code
- is integrated in established development frameworks such as eclipse with Junit and Ant and service-oriented computing technologies such as Web Services and Semantic Web technologies

The major advantages of my rule-based approach are:

- it allows to continuously adapt SLAs to a rapidly changing business environment, and overcomes the restricting nature of slow change cycles
- it lowers the cost incurred in the modification of contractual / business logic
- it shortens development time of SLA specifications and safeguards the engineering process
- rules are externalized and easily shared among multiple applications and contractual partners
- changes can be made faster and with less risk
- it provides are more human way to describe, manage and maintain formalized SLAs as compared to implicit procedural implementations
- it provides highly reliable and traceable results derived by generic inference engines with logical semantics based on a solid mathematical basis
- it allows to express and declaratively program highly sophisticated SLA logic and business management policies

## 3. Problem Solution

Based on a functional and non-functional requirement analysis on real-world SLAs, commercial SLA management tools and SLA languages as well as general adequacy criteria for KR languages in artificial intelli-gence (epistemological, algorithmic, logic-formal, ergonomical adequacy) in particular with respect to expressiveness and computational complexity I have evaluated different logical formalisms and **combined selected logics in one expressive KR framework called ContractLog.** ContractLog basically uses normal logic programs with derivation rules extended with default negation (negation-as-failure) and explicit negation and a declarative semantics (extended well-founded semantics). The KR implements further logical formalisms which are needed to fully represent formal SLA specifications. Table 1 summarizes the main concepts used in ContractLog.

**Table 1: Core Logic Concepts implemented in ContractLog KR**

| Logic | Formalism | Usage |
|---|---|---|
| Extended Logic | Derivation Rules | Deductive reasoning on SLA rules extended with negation-as-finite-failure and explicit negation. |
| Typed Logic | Object-oriented Typed Logic and Procedural Attachments | Typed terms restrict the search space and enable object-oriented software engineering principles. Procedural attachments integrate object oriented programming into declarative rules. → integration of external systems |
| Description Logic | Hybrid Description Logic Types and Semantic Web ontology languages | Semantic domain descriptions (e.g., contract ontologies) in order to describe rules domain independent. → external contract/domain vocabularies |
| (Re)active Logic | Event-Condition-Action Rules (ECA) | Active event detection/event processing and event-triggered actions. → reactive rules. |
| Temporal Event/Action Logic | Event Calculus | Temporal reasoning about dynamic systems, e.g. interval-based complex event definitions (event algebra) or effects of events on the contract state → contract state tracking<br>→ reasoning about events/actions and their effects |
| Deontic Logic | Deontic Logic with norm violations and exceptions (section 3.6) | Rights and obligations formalized as deontic contract norms with norm violations (contrary-to-duty obligations) and exceptions (conditional. defeasible obligations). → normative deontic rules. |
| Integrity Preserving, Preferenced, Defeasible Logic | Defeasible Logic and Integrity Constraints (section 3.7) | Default rules and priority relations of rules. Facilitates conflict detection and resolution as well as revision/updating and modularity of rules. → default rules and rule priorities |
| Test Logic | Test-driven Verification and Validation for Rule Bases | Validation and Verification of SLA specifications against predefined SLA requirements<br>→ safeguards the engineering, dynamic adaption and interchange process of SLAs |

I have implemented an extended goal-driven linear resolution[1] for computing well-founded semantics with a typed unification, support for procedural attachments and a unitized knowledge base (KB) where rule sets are managed as ID-based modules. This procedural semantics enables the tight integration of Java functionalities and Semantic Web ontologies into rule resolution and facilitates the dynamic evolution of the KB. To implement the different (non-monotonic) logics in ContractLog I use meta-programming techniques, where the formalisms are also expressed as a logic program in terms of derivation rules. The meta programs are managed as a library of script files (stand-alone modules), which can be dynamically imported on a "per-need-basis". Meta-programming and meta-interpreters are a popular technique in logic programming [1]. My ECA interpreter, which is implemented as generalized add-on for arbitrary backward-reasoning inference engines, simulates the forward-directed operational semantics of ECA rules via frequently querying the knowledge base and hence enables active rules in combination with derivation rules in LPs. It supports parallel ECA rule execution, transactional (bulk) updates and in general complex event processing using my novel interval-based event calculus variant as an event algebra. ContractLog comprises many more implementations which I can not mention here. More details can be found in on the project site [2] and the papers and reports about the KR.

---

[1] The linearity of my novel resolution with goal memoization and loop prevention for WFS enables efficient stack based memory structures and strictly sequential operators to be used, in contrast to other procedural tabling based semantics such as SLG resolution.

ContractLog is extended by a declarative **r**ule **b**ased **SLA** Markup Language (**RBSLA**) which is based on the emerging Semantic Web Rule standard RuleML [3] in order to provide a compact and user-friendly SLA related serialization syntax which facilitates rule interchange in a distributed environment, serialization in XML and tool support. Therefore, it adapts and extends RuleML to the needs of the SLA domain. RBSLA is designed with a layered structure where each layer adds different modelling expressiveness to the RuleML core.

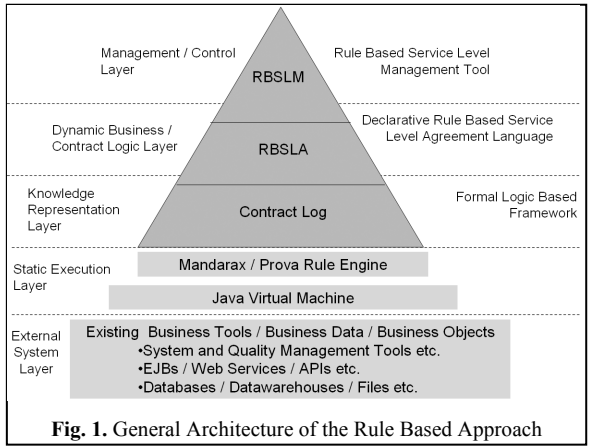

**Fig. 1.** General Architecture of the Rule Based Approach

Based on ContractLog and the higher-level RBSLA we have implemented a **r**ule **b**ased **s**ervice **l**evel **m**anagement (**RBSLM**) prototype as a test bed and proof of concept implementation. The RBSLM tool splits into the **Contract Manager** (CM) and the **Service Dashboard** (SD). The CM is used to manage, write, maintain and update SLA rules with support for different roles. The SD visualizes status information and metrics during the monitoring and execution process. Figure 1 illustrates the general architecture of our rule based SLA management approach in a layered model.

## 4. Research Methodology and Evaluation

My work follows a constructivist, SE-oriented methodology and presents a proof-of-concept implementation. I adopt the Design Science Research approach, as described e.g. in Hevner et al. [4] and **propose a new design artifact** which defines an expressive, declarative, logic-based KR and a SLA mark-up language for representation and automation of SLAs based on logic programming techniques and selected further adequate KR formalisms. With my **alternative approach I provide new levels of flexibility and automation** which are not available in the current technologies and tools in the SLA domain. I try to **overcome real-world problems which are of high relevance and importance** for SLA representation such as rapidly changing, highly-distributed and loosely coupled service oriented environments, slow change cycles, and new business models (on-deman, utility computing) as well as several new regulations and laws with compliance rules (Saban Oxley, Basel II etc.). I have evaluated the novel rule based SLA representation approach and the ContractLog KR w.r.t. established methodologies in SE, KR and LP in several ways, via *theoretical worst-case analysis*, *experimentally* via performance *benchmark tests* and formalized real-world SLA *simulations* illustrating expressiveness as well as on *use cases* which have been submitted to *open-source projects* (Mandarax, Prova, RBSLA) and *standardization initiatives* (RuleML, W3C Rule Interchange Format RIF Use Cases). Although, the worst-case complexity analysis for full expressiveness in the ContractLog KR reveals relatively high complexity bounds, e.g. NEXPTIME for OWL-DL-typed, non-recursive logic programs with unrestricted functions, it achieves much better results in practice, because typical SLA formalization do not need the full expressiveness and complex computations (from a logical perspective) such as mathematical functions or complex data selection queries can be shifted to highly optimized, external code and query languages, e.g. Java code or SQL queries. The benchmark experiments evaluate performance of query answering in different logic program classes (e.g. propositional, datalog, normal) with different scalable problem sizes in order to measure performance of crucial inference properties such as rule chaining, recursion, variable unification, goal memoization, defeasible reasoning, temporal reasoning etc. Various metrics such as number of facts, number of rules, overall size of literals indicating the size of complexity of a particular benchmark test might be used to estimate the time for query answering and memory consumption. The experiments reveal adequate performance and high scalability of the ContractLog KR even for larger problem sizes with thousands of rules. This proves computational adequacy of the proposed rule-based SLA representation approach under real-world settings. In fact, I have formalized typical real-world SLAs from different industries in ContractLog within several dozens up to hundreds rules (see e.g. RIF use cases[2]) which can be efficiently executed and monitored within milliseconds. The analyses and qualitative comparisons with other representation approaches such as WSLA,

---

[2] http://www.w3.org/2005/rules/wg/wiki/Rule_Based_Service_ Level_Management_and_SLAs_for_Service_Oriented_Computing

which result from the implemented use cases and case studies, have revealed the higher flexibility and automation of our rule based approach.

In summary, our work fulfils both primary demands of Design Science, namely **relevance** and **rigor** and there is large evidence that the basic research question, whether KR techniques and logic programming can be used to adequately represent and automate SLAs, can be answered positively.

## 6. Conclusion

In this proposal I have described a declarative rule based approach to automated SLA representation and management. I have given an insight into the logical core, the ContractLog framework which underpins the declarative RBSLA language. In contrast to conventional pure procedural programming approaches my declarative logic based approach simplifies maintenance, management and execution of SLA rules and allows easy combination and revision of rule sets to build sophisticated and graduated contract agreements, which are more suitable in a dynamic service oriented environment than the actually used, simplified rules.

The intended final submission of my thesis is November/December 2006. Hence, the PhD symposium will be a good opportunity to present my complete proposal and the achieved results, discuss them with the audience and use the feedback to sharpen my argumentation line for the final defence and the refinement of my thesis.

**Links:**

RBSLA project: http://ibis.in.tum.de/staff/paschke/rbsla/index.htm
RBSLA distribution: https://sourceforge.net/projects/rbsla